\documentclass[a4paper,fleqn]{cas-sc}

\usepackage[numbers]{natbib}
\usepackage{adjustbox}
\usepackage{rotating}
\usepackage{placeins}
\usepackage{lineno}
\usepackage{algorithm}
\usepackage{algorithmic}
\usepackage{amsmath}
\usepackage[flushleft]{threeparttable}
\usepackage{subcaption}
\usepackage{multirow}

\def\tsc#1{\csdef{#1}{\textsc{\lowercase{#1}}\xspace}}
\tsc{WGM}
\tsc{QE}
\tsc{EP}
\tsc{PMS}
\tsc{BEC}
\tsc{DE}

\begin{document}
\let\WriteBookmarks\relax
\def\floatpagepagefraction{1}
\def\textpagefraction{.001}
\shorttitle{MobiCLR: Mobility Time Series Contrastive Learning for Urban Region Representations}

\shortauthors{N Kim \textit{et al.}}

\title [mode = title]{MobiCLR: Mobility Time Series Contrastive Learning for Urban Region Representations}                      

%
\author[1]{Namwoo Kim}
\fnmark[1]
\ead{ih736x@kaist.ac.kr}
\credit{Conceptualization of this study, Methodology, Software}
\affiliation[1]{organization={Department of Civil and Environmental Engineering, Korea Advanced Institute of Science and Technology},
    city={Daejeon},
    postcode={34141}, 
    country={South Korea}}

\affiliation[2]{organization={Department of Technology Management and Innovation and the Center for Urban Science and6
Progress (CUSP), New York University},
    city={Brooklyn},
    postcode={11201}, 
    state={NY},
    country={United States}}
\affiliation[3]{organization={Department of Industrial and Systems Engineering, Korea Advanced Institute of Science and Technology},
    city={Daejeon},
    postcode={34141}, 
    country={South Korea}}

\author%
[2]
{Takahiro Yabe}
\ead{ty2520@nyu.edu}

\author%
[3]
{Chanyoung Park}
\ead{cy.park@kaist.ac.kr}

\author%
[1]
{Yoonjin Yoon}
\cormark[1]
\ead{yoonjin@kaist.ac.kr}

\cortext[cor1]{Corresponding author (Yoonjin Yoon: yoonjin@kaist.ac.kr)}

\fntext[fn1]{Namwoo Kim is the first author.}

\begin{abstract}
Recently, learning effective representations of urban regions has gained significant attention as a key approach to understanding urban dynamics and advancing smarter cities. Existing approaches have demonstrated the potential of leveraging mobility data to generate latent representations, providing valuable insights into the intrinsic characteristics of urban areas. However, incorporating the temporal dynamics and detailed semantics inherent in human mobility patterns remains underexplored. To address this gap, we propose a novel urban region representation learning model, \textit{Mobility Time Series Contrastive Learning for Urban Region Representations (MobiCLR)}, designed to capture semantically meaningful embeddings from inflow and outflow mobility patterns. MobiCLR uses contrastive learning to enhance the discriminative power of its representations, applying an instance-wise contrastive loss to capture distinct flow-specific characteristics. Additionally, we develop a regularizer to align output features with these flow-specific representations, enabling a more comprehensive understanding of mobility dynamics. To validate our model, we conduct extensive experiments in Chicago, New York, and Washington, D.C. to predict income, educational attainment, and social vulnerability. The results demonstrate that our model outperforms state-of-the-art models.

\end{abstract}

\begin{keywords}
urban region embedding\sep time series contrastive learning\sep human mobility\sep urban computing
\end{keywords}

\begin{titlepage}
    \begin{center}

        \LARGE
        \textbf{MobiCLR: Mobility Time Series Contrastive Learning for Urban Region Representations}
    
            
       \vspace{1.5cm}

        \Large 
       \textbf{Namwoo Kim$^1$, Takahiro Yabe$^2$, Chanyoung Park$^3$, and Yoonjin Yoon$^1$}
        
        \large
       \vspace{1.5cm}
       $^1$Department of Civil and Environmental Engineering, Korea Advanced Institute of Science and Technology (KAIST)
       
       \vspace{0.2cm}
       $^2$Department of Technology Management and Innovation and the Center for Urban Science and Progress (CUSP), New York University
       
       \vspace{0.2cm}
       $^3$Department of Industrial and Systems Engineering, Korea Advanced Institute of Science and Technology (KAIST)
    \end{center}

    \small
    \vfill

    \noindent \textbf{Author}: Namwoo Kim, Ph.D
    
    \noindent \textit{email}: ih736x@kaist.ac.kr

    \vspace{0.2cm}

    \noindent \textbf{Author}: Takahiro Yabe, Ph.D

    \noindent \textit{email}: ty2520@nyu.edu
    
    \vspace{0.2cm}
    \noindent \textbf{Author}: Chanyoung Park, Ph.D
    
    \noindent \textit{email}: cy.park@kaist.ac.kr

    \vspace{0.2cm}
    
    \noindent \textbf{Corresponding author}: Yoonjin Yoon, Ph.D
    
    \noindent \textit{email}: yoonjin@kaist.ac.kr
    
    \noindent \textit{phone}: +82) 042-350-3615
    
    \noindent \textit{address}: W16 410, Department of Civil and Environmental Engineering, Korea Advanced Institute of Science and Technology, Daejeon, 34141, South Korea
    
    
    
\end{titlepage}

\maketitle

\section{Introduction}
Urban regions are dynamic and complex spaces where diverse social and economic structures interact with human activity. Comprehending this intricate relationship is imperative for fostering the development of livable and sustainable cities \cite{zhang2021unveiling}. In recent years, utilizing various urban data to learn the latent representations of urban regions has gained significant attention in the field of urban computing \cite{fu2019efficient, liang2023revealing, li2023urban, zhao2023learning}. Urban region representation learning involves transforming the diverse attributes of urban regions into a latent vector space, facilitating a deeper understanding of urban dynamics. The embeddings derived from this process have proven valuable for a range of applications, such as land-use classification \cite{zhang2021multi}, socio-economic and demographic indicator prediction \cite{huang2021m3g}, and crime prediction \cite{wang2017region}.

With advancements in sensing technology, the use of mobility data—such as taxi trip records—has proven effective in learning semantically meaningful embeddings of urban regions. Numerous studies have attempted to use trip record data to learn region embeddings \cite{wang2017region, yao2018representing, kim2022effective, zhang2021multi, li2024urban, zhang2022region}. These studies can be divided into three categories: sequence modeling, OD (origin-destination) matrix reconstruction, and contrastive learning. The sequence modeling approach includes methods that capture OD co-occurrences to learn region representations \cite{yao2018representing} or generate random-walk-based sequences on spatial and flow graphs using mobility flow data \cite{wang2017region}. The OD matrix reconstruction method involves reconstructing conditional trip distributions based on the influx and outflux of human mobility from a region over a specified period \cite{kim2022effective, zhang2021multi}. Lastly, the contrastive learning approach generates influx and outflux embeddings from trip record data and applies a contrastive loss to obtain the region representations \cite{zhang2022region, li2024urban}. 

While previous methods have demonstrated effectiveness, they exhibit limitations in two key aspects. First, they often fall short in fully leveraging the rich temporal dynamics embedded in mobility data. In particular, the temporal patterns of inbound and outbound flows offer valuable insights into the functional dynamics of urban regions. For instance, regions with high morning outflux and evening influx typically serve as residential areas, while increased weekend influx often signals entertainment districts. By incorporating such information, we could gain deeper understanding of how urban regions function and operate over time. Second, the use of region embedding models for inferring and predicting composite indices that capture the multifaceted nature of urban environments has been limited. Most existing approaches focus on single indicators, such as income or educational attainment, which only reveal specific aspects of a region. Such single-dimensional analysis fails to account for the complex interactions between multiple social, economic, and environmental factors that define urban dynamics.

In this paper, we proposed a novel region embedding model called \textit{mobility time series contrastive learning for region representations} (MobiCLR). To better capture dynamic inflow and outflow patterns within urban regions, our model incorporated time series representation learning. Using two weeks of ride-hailing and taxi-trip records, we generated hourly time-series data on inbound and outbound trips across urban regions. Contrastive learning was employed to enhance the discriminative power of the learned representations. To effectively extract useful features from mobility patterns, our model leverages instance-wise contrastive loss to capture inbound- and outbound-specific characteristics. Additionally, we designed a novel regularizer to align output features with these specific patterns, enabling the model to capture comprehensive mobility information. For validation, we utilized state-of-the-art models to evaluate the proposed approach in predicting both single indicators (income and educational attainment) as well as composite index (social vulnerability index). Through extensive experiments on data from Chicago, New York, and Washington D.C., MobiCLR was shown to outperform all baseline models in predicting both single and composite index.

Our main contributions are highlighted as follows:
\begin{itemize}

\item We proposed a novel mobility time series contrastive learning framework to capture the temporal dynamics of mobility patterns. Specifically, we incorporated an instance-wise contrastive objective to learn distinct representations for inbound and outbound flows, enhancing the model’s ability to capture flow-specific characteristics. Additionally, we designed an auxiliary regularizer to align these representations, resulting in a comprehensive view of regional mobility dynamics.

\item We conducted extensive experiments in Chicago, New York, and Washington D.C., to evaluate the performance of the proposed framework in predicting single indicators (income and educational attainment) as well as composite index (social vulnerability index). The experimental results demonstrate significant performance improvements over state-of-the-art models in predicting both single and composite indicators. 
\end{itemize}

The remainder of this paper is organized as follows. Section 2 presents a literature review on knowledge extraction using mobility data and time series contrastive learning. Section 3 defines the problem at hand. The proposed framework is described in Section 4. Section 5 describes the experimental setup, datasets, baselines, and prediction results. Section 6 presents additional analysis, including the choice of data augmentation strategies, ablation studies, and transferability tests. Finally, Section 7 presents the conclusions.

\section{Related Work}
\subsection{Urban Region Embedding}
Human mobility data has become a valuable resource for gaining insights into various aspects of urban dynamics, and numerous studies have integrated it to improve our understanding of urban environments \cite{carroll2021community, chi2022microestimates, aiken2022machine, xia2023assessing, liu2021urban, liu2024exploring, gao2020semantic}. Mobility data provides valuable insights into urban flow patterns \cite{liu2021urban, chen2021multiple}, hazard exposure \cite{carroll2021community, xia2023assessing}, region recommendation \cite{liu2024exploring}, and fine-grained socio-economic assessments \cite{chi2022microestimates, aiken2022machine}. Recently, several studies have combined mobility data with deep learning techniques to deepen this understanding, focusing on three main approaches: sequence modeling \cite{yao2018representing, wang2017region}, OD matrix reconstruction \cite{kim2022effective, zhang2021multi, wu2022multi}, and contrastive learning \cite{li2024urban, zhang2022region}.

In sequence modeling methods, region embeddings are extracted by treating individual vehicle origin-destination trip chains as sequences and modeling the co-occurrence of origin and destination regions to learn region representations. \cite{yao2018representing}. Alternatively, inflow and outflow data from taxi trip records are used to weight the edges of a geo-spatial graph. A random walk on this weighted graph generates sequences that, combined with the skip-gram objective, produce region embeddings \cite{wang2017region}. In OD matrix reconstruction methods, the OD pair and trip volume information are used to model inter-region interactions as a conditional trip distribution \cite{kim2022effective, zhang2021multi}. Lastly, in the contrastive learning approach, inflow and outflow representations are averaged to obtain overall region representations, followed by the application of contrastive loss \cite{li2024urban, zhang2022region}. 

While these advancements have shown promising results, they remain limited in capturing the full temporal dynamics and semantic richness of human mobility patterns. By considering both temporal dynamics and the interactions between inbound and outbound mobility, we could gain deeper insights into regional functions and a more nuanced understanding of how regions operate over time.

\subsection{Contrastive Time Series Representation Learning}
 Supervised learning relies heavily on labeled data, which can be challenging and expensive to obtain in real-world scenarios. To overcome these limitations, contrastive learning has emerged as a promising approach for learning intrinsic patterns present in data without relying on external annotations or labels. It uses positive or negative pairs of data to learn representations. Through data augmentation, the positive pairs comprise two augmented versions of the same input data. In contrast, negative pairs are formed using different input samples. During the training process, the model maps input samples into the latent vector space, aiming to bring positive pairs geometrically closer, while the negative pairs are pushed further apart. 
 
Building on this foundation, contrastive learning techniques for time series data have shown promising results. For example, Franceschi et al. \cite{franceschi2019unsupervised} employed time-based negative sampling to encourage the model to learn semantics similar to those of the smapled sub-series. Eldele et al. \cite{tstcc} employed augmentation techniques, including scaling and permutation, to extract representations invariant to transformations. Tonekaboni et al. \cite{tonekaboni2020unsupervised} considered the temporal dependency of time series data by ensuring that neighboring timesteps are distinguishable from those that are not adjacent. Yue et al. \cite{Yue2021TS2VecTU} used hierarchical contrastive loss to extract a contextually invariant representation. Wickstr$\phi$m et al. \cite{wickstrom2022mixing} proposed a novel augmentation strategy to predict the mixing proportion of time series samples. Meng et al. \cite{meng2023mhccl} employed hierarchical clustering to construct contrastive pairs, thereby preventing the inclusion of instances with semantics similar to false-negative pairs.

\section{Definition and Problem Statement}
In this Section, we provide a formal definition for the mobility time series and present the problem statement.

\subsection{Inbound and Outbound Time Series}
The hourly counts of inbound and outbound trips for each location $n$ were computed from the collected mobility data. This yields, the time series $x_{n}^i=\{x_{n,t}^i\}_{t=1}^T\in\mathbb{R}^{T\times1}$ and $x_n^o=\{x_{n,t}^o\}_{t=1}^T\in\mathbb{R}^{T\times1}$ for the inbound and outbound trips for $\forall n\in\{1,2, \ldots, N\}$, respectively. Here, $T$ denotes a given time interval, and $N$ represents the number of areas of interests. Consequently, we obtain the mobility time series $x^{io}_n=[x_{n}^i, x_{n}^o]\in\mathbb{R}^{T\times2}$ by concatenating those two time series $x_n^i$ and $x_n^o$. 

\subsection{Problem Statement}
For a set of mobility time series $\mathcal{X}=\{x^{io}_{1},x^{io}_{2},\ldots,x^{io}_{n},\ldots,x^{io}_{N}\}$ of $N$ regions, the objective is to train a neural network $f_{\theta}$ that maps each $x^{io}_{n}$ to their corresponding representation $h_{n}\in\mathbb{R}^{D}$, where $D$ is the dimension of the latent vector space. 
\section{Methodology}
In this Section, we introduce the MobiCLR framework. We first present an overview of the proposed model, followed by a explanation of the steps involved in learning region representations. 

\subsection{Model Architecture}
\begin{figure*}[h]
\includegraphics[width=0.9\textwidth]{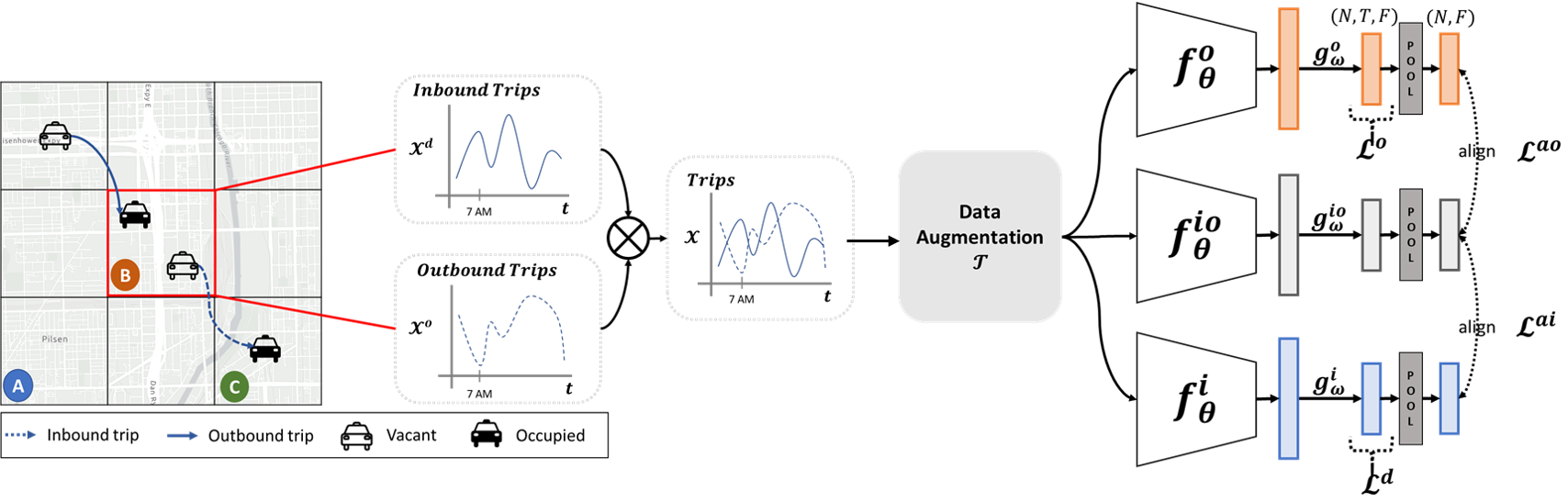}
\centering
\caption{Overall framework of MobiCLR. We first pre-process region-wise hourly inbound and outbound trips. Subsequently, a two-step data augmentation process is applied to the mobility time series data, followed by the feeding of this augmented data into three encoders: $f_\theta^i$, $f_\theta^o$, and $f_\theta^io$. Here, $f_\theta^i$ and $f_\theta^o$ learn inbound- and outbound-specific human mobility patterns in each region, respectively, while $f_\theta^{io}$ extracts semantics containing both inbound- and outbound-specific features. After pre-training, the embedding vectors obtained through $f_\theta^{io}$ are utilized in downstream applications.}
\label{fig:framework}
\end{figure*}

The overall framework of MobiCLR is shown in Figure \ref{fig:framework}. The proposed framework involved the application of data augmentation to the original time series, followed by the extraction of transformation-invariant features using time series encoders. Subsequently, the model was trained to learn contextual information regarding urban regions related to both the origin and destination characteristics of human mobility.  

The proposed model employs three types of encoders. The encoders $f_\theta^i$, $f_\theta^o$, and $f_\theta^{io}$ take $x^i$, $x^o$, and $x^{io}$ as inputs, respectively. The encoders $f_\theta^o$ and $f_\theta^i$ capture the temporal context of the time series of outbound/inbound trips, respectively. On the other hand, $f_\theta^{io}$ extracts overall semantics that encompass both inbound- and outbound-specific features. Thus $f_\theta^{io}$ aims to capture the characteristics of both inbound and outbound activities, allowing for a more holistic representation of the data. Consequently, we obtained three representation vectors: $h_n^i\in\mathbb{R}^{T\times D}$, $h_n^o\in\mathbb{R}^{T\times D}$, and $h_n^{io}\in\mathbb{R}^{T\times D}$ from each of the encoder. 

Each of the time series encoders comprised an input projection layer and dilated convolutional neural network module \cite{Fisher, oord2016wavenet}. A stack of three convolution blocks with skip connections between adjacent blocks was utilized to capture the temporal dependencies of the time series data.

\subsection{Data Augmentation}
\begin{figure}[!h]
\centering
     \includegraphics[width=0.7\textwidth]{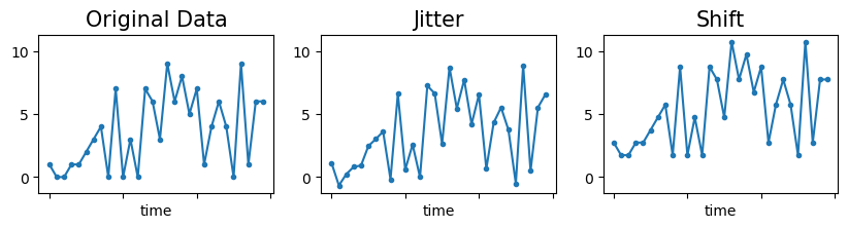}
    \caption{Graphical illustration of data augmentation strategy.}
    \label{fig:aug_illustration}
\end{figure}

Data augmentation techniques were utilized in conjunction with contrastive learning to extract meaningful semantics from mobility time series data. In this study, a two-step data augmentation process involving jittering and shifting was implemented. The resulting representations from the encoder $f_\theta$ of the timestamp $t$ from two augmentations of $x_n$ are represented as $h_{n,t}$ and $\widetilde{h}_{n,t}$. The augmentation strategy is explained as follows. 
\begin{itemize}\item \textbf{Jitter:} Independent and identically distributed Gaussian noise is added to each time step, sampled from a Gaussian distribution with a mean of 0 and a standard deviation of 0.2 (i.e., $\epsilon_t\sim N(0,\ 0.2)$). Each time step is now jittered as ${\widetilde{x}}_t={\epsilon_t}\times x_t$.
\item \textbf{Shift:} The time series data was shifted by a single random scalar value, obtained by sampling from a Gaussian distribution with a mean of 0 and a standard deviation of 0.2 (i.e., $\epsilon\sim N(0,\ 0.2)$). Each time step was then shifted as ${\widetilde{x}}_t={\epsilon+x}_t$.
\end{itemize}
 To enhance clarity, a graphical illustration of augmentation strategy is provided in Figure \ref{fig:aug_illustration}. 

\subsection{Learning Objectives}

To extract information embedded within mobility patterns, we utilize instance-level contrastive loss and auxiliary regularizer. Through instance-level contrastive loss, the model learns flow-specific characteristics, while the auxiliary regularizer enables it to capture comprehensive regional mobility patterns.


\subsubsection{Instance-level contrastive loss}
To apply contrastive loss, the projection headers $g_\omega^i$, $g_\omega^o$, and $g_\omega^{io}$ were employed on each of the representation vectors $h_n^i$, $h_n^o$, and $h_n^{io}$, respectively. This yielded $z_n^i\in\mathbb{R}^{T\times F}$, $z_n^o\in\mathbb{R}^{T\times F}$, and $z_n^{io}\in\mathbb{R}^{T\times F}$ for $\forall n\in\{1,2,\ldots,\ N\}$.

An instance-level contrastive loss was applied to capture the distinct semantics of inbound and outbound trips within regions. From the data augmentation, the embeddings $z_n^i$, ${\widetilde{z}}_n^i$, $z_n^o$, and ${\widetilde{z}}_n^o$ were obtained, which were then used to compute the inbound-specific contrastive loss $\ell^i$ and outbound-specific contrastive loss $\ell^o$ by leveraging the normalized temperature-scaled cross entropy (NT-Xent) loss \cite{sohn2016improved, wu2018unsupervised, oord2018representation}. Let $sim\left(a,b\right)$ denote the cosine similarity between vectors $a$ and $b$, and let $\tau$ be the temperature hyperparameter. We denote the similarity between the representations $z_{n,t}$ and $z_{m,t}$ as $\exp\left(sim\left(z_{n,t},z_{m,t}\right)/\tau\right)$.
The contrastive losses for positive pairs $(z_{n,t}^i,{\widetilde{z}}_{n,t}^i)$ and $(z_{n,t}^o,{\widetilde{z}}_{n,t}^o)$ are formulated respectively as:

\begin{equation}
\begin{split}
{}& \ell^i\left(z_{n,t}^i,{\widetilde{z}}_{n,t}^i\right) = \\
& -log\frac{
\exp\left(sim\left(z_{n,t}^i,{\widetilde{z}}_{n,t}^i\right)/\tau\right)
}{\sum\limits_{m\in B}{\mathbf{1}_{\left[m\neq n\right]}
\exp\left(sim\left(z_{n,t}^i,z_{m,t}^i\right)/\tau\right)}+\sum\limits_{m\in B} 
\exp\left(sim\left(z_{n,t}^i,{\widetilde{z}}_{m,t}^i\right)/\tau\right)}
\end{split}
\end{equation}
\begin{equation}
\begin{split}
{}& \ell^o\left(z_{n,t}^o,{\widetilde{z}}_{n,t}^o\right) = \\
& -log\frac{
\exp\left(sim\left(z_{n,t}^o,{\widetilde{z}}_{n,t}^o\right)/\tau\right)
}{\sum\limits_{m\in B}{\mathbf{1}_{\left[m\neq n\right]}
\exp\left(sim\left(z_{n,t}^o,z_{m,t}^o\right)/\tau\right)}+\sum\limits_{m\in B} 
\exp\left(sim\left(z_{n,t}^o,{\widetilde{z}}_{m,t}^o\right)/\tau\right)}
\end{split}
\end{equation}
where $\mathcal{B}$ is a batch size and $\mathbf{1}_{\left[m\neq n\right]}$ is an indicator function that is 1 if $m\neq n$ and 0 otherwise. Consequently, the inbound-specific contrastive loss $\mathcal{L}^i$ and outbound-specific contrastive loss $\mathcal{L}^o$ for a minibatch are obtained as
\begin{equation}
\mathcal{L}^i=\frac{1}{\left|\mathcal{B}\right|\times T}\sum_{n\in B}\sum_{t=1}^{T}{\ell^i\left(z_{n,t}^i,{\widetilde{z}}_{n,t}^i\right)}
\end{equation}
\begin{equation}
\mathcal{L}^o=\frac{1}{\left|\mathcal{B}\right|\times T}\sum_{n\in B}\sum_{t=1}^{T}{\ell^o\left(z_{n,t}^o,{\widetilde{z}}_{n,t}^o\right)}
\end{equation}

\subsubsection{Auxiliary regularizer}
An auxiliary regularizer was proposed to improve representation learning by aligning output features with inbound- and outbound-specific characteristics. This regularizer bridges these distinct representations, combining inbound and outbound dynamics into a unified and comprehensive view of regional mobility patterns. In contrast to Eqs. (2) and (3), the approach utilizes average pooling along the temporal dimensions of the $z_n^i$, $z_n^o$, and $z_n^{io}$ to generate summary vectors, $z_{n,\ast}^i \in\mathbb{R}^{F}$, $z_{n,\ast}^o\in\mathbb{R}^{F}$, and $z_{n,\ast}^{io}\in\mathbb{R}^{F}$, respectively. Subsequently, contrastive loss was applied to maximize the agreement between $z_{n,\ast}^i$ and $z_{n,\ast}^{io}$, as well as between $z_{n,\ast}^o$ and $z_{n,\ast}^{io}$.

The contrastive losses for the positive pairs $(z_{n,\ast}^{io},z_{n,\ast}^i)$ and $(z_{n,\ast}^{io},z_{n,\ast}^o)$ are defined as:

\begin{equation}
\begin{split}
    {}&\ell^{ai}\ \left(z_{n,\ast}^{io},z_{n,\ast}^i\right)=\\&-log\frac{D_{\tau_a}\left(z_{n,\ast}^{io},z_{n,\ast}^i\right)}{\sum\limits_{m\in B}{\mathbf{1}_{\left[m\neq n\right]}D_{\tau_a}\left(z_{n,\ast}^{io},z_{m,\ast}^{io}\right)}+\sum\limits_{m\in B} D_{\tau_a}\left(z_{n,\ast}^{io},z_{m,\ast}^i\right)}
\end{split}
\end{equation}

\begin{equation}
\begin{split}
{}&\ell^{ao}\left(z_{n,\ast}^{io},z_{n,\ast}^o\right)=\\&-log\frac{D_{\tau_a}\left(z_{n,\ast}^{io},z_{n,\ast}^o\right)}{\sum\limits_{m\in B}{\mathbf{1}_{\left[m\neq n\right]}D_{\tau_a}\left(z_{n,\ast}^{io},z_{m,\ast}^{io}\right)}+\sum\limits_{m\in B} D_{\tau_a}\left(z_{n,\ast}^{io},z_{m,\ast}^o\right)}
\end{split}
\end{equation}
Then, the auxiliary regularizers for a mini-batch are computed as:
\begin{equation}
\begin{split}
\mathcal{L}^{a}=\\&\frac{1}{\left|\mathcal{B}\right|}\sum_{n\in B}[{\ell}^{ai}\ \left(z_{n,\ast}^{io},z_{n,\ast}^i\right)+\ell^{ai}\ \left({\widetilde{z}}_{n,\ast}^{io},{\widetilde{z}}_{n,\ast}^i\right)] + \\& \frac{1}{\left|\mathcal{B}\right|}\sum_{n\in B}[{\ell}^{ao}\ \left(z_{n,\ast}^{io},z_{n,\ast}^o\right)+\ell^{ao}\ \left({\widetilde{z}}_{n,\ast}^{io},{\widetilde{z}}_{n,\ast}^o\right)]
\end{split}
\end{equation}

Incorporating regularizers into the model enhances its learning capacity by exposing it to a wide range of positive and negative sample variations. Consequently, the model generates more robust and informative representations.

\subsection{Overall objective}
Finally, the overall objective function $\mathcal{L}$ is obtained during training as the combination of $\mathcal{L}^i$,$\mathcal{L}^o$, and $\mathcal{L}^{a}$, where $\mathcal{L}^i$ and $\mathcal{L}^o$ enforce the learning of inbound- and outbound-specific features, respectively, while $\mathcal{L}^{a}$ aligns output features with inbound- and outbound-specific features. The overall objective function $\mathcal{L}$ is formulated as follows:
\begin{equation}
\mathcal{L}=\mathcal{L}^i+\mathcal{L}^o+{\mathcal{L}}^{a}
\end{equation}

\section{Experiments}

This Section describes the datasets used in the experiment, as well as the baseline models, experimental setup, and performance comparison. 

\subsection{Experimental Setup}
The entire codebase was implemented in PyTorch \cite{paszke2019pytorch}. We trained the model exclusively on the training set, and the pre-trained model was applied to the test set to obtain representations. Specifically, 75\% of the data were used for training, while the remaining 25\% were used for testing. Results were reported as the average of 5 runs. The default batch size was set to 4, and the learning rate was fixed at 0.0001. The model was trained for 30 epochs and the representation dimension was set to 128. The kernel size for all one-dimensional convolution layers was set to 3, and each hidden dilated convolution had a channel size of 128. The temperature parameters $\tau$ and $\tau_a$ were set to 1 and 0.1, respectively. All the experiments were conducted on an NVIDIA GeForce RTX 3090GPU.

\subsection{Datasets}
We utilized taxi trip record data to extract region embeddings and evaluated the proposed model by predicting single indicators, including educational attainment and income, as well as a composite index, specifically the Social Vulnerability Index. Two weeks of taxi trip data were collected across three major US cities: New York, Washington D.C, and Chicago. The taxi trip data were acquired from the open data portal of each city\footnote[1]{https://data.cityofchicago.org/} \footnote[2]{ https://opendata.cityofnewyork.us/} \footnote[3]{ https://opendata.dc.gov/}. The statistics of the mobility time series data are listed in Table \ref{tab:stat}. For single indicator prediction, educational attainment and per capita income data were retrieved from the American Community Survey (ACS) API \footnote[4]{https://www.census.gov/en.html}. For the composite index prediction, social vulnerability index was used. The social vulnerability index employs a 16-factor such as unemployment, health insurance, age, and vehicle access. Social vulnerability index for the three cities were sourced from the Centers for Disease Control and Prevention (CDC) \footnote[5]{https://www.atsdr.cdc.gov/}. A detailed description of the dataset is provided below:
\begin{itemize}
\item \textbf{Taxi trip record} The mobility data for Chicago, New York, and Washington D.C. were collected through each city's data portal. Each dataset included the start and end times of each trip, pickup and drop-off locations in the community area ID, distance traveled, fare amount, tip amount, etc. To generate a mobility time series, a spatial join operation was performed to match the origin and destination location.

\item \textbf{Educational Attainment} College graduation rates (scaled from 0 to 1) were obtained from the ACS 5-Year data.

\item \textbf{Income} Data on per capita income was obtained from the ACS 5-Year data.

\item \textbf{Social vulnerability index} Social vulnerability is defined as the susceptibility of individuals, communities, or populations to adverse impacts resulting from natural hazards or external stressors \cite{cutter2003social, bankoff2004mapping, adger2006vulnerability}. In this study, we used the social vulnerability index provided by the CDC as the ground truth for regional social vulnerability. The social vulnerability index employs a 16-factor approach to rank places on social factors, including unemployment, minority status, disability, etc., and aggregates the rankings into a single scale ranging as 0-1 (least vulnerable to most vulnerable).

\end{itemize}

\begin{table}[]
\centering
\caption{Statistics of mobility time series data of three US cities: Chicago (CH), New York (NY), Washington DC (DC).}
\label{tab:stat}

\begin{tabular}{cccc}
\hline
                    & CH                                                               & NY                                                              & DC                                                              \\ \hline
\# Regions          & 77                                                               & 279                                                             & 178                                                             \\
Periods             & \begin{tabular}[c]{@{}c@{}}04/01 $\sim$15 \\ (2016)\end{tabular} & \begin{tabular}[c]{@{}c@{}}04/01 $\sim$15\\ (2016)\end{tabular} & \begin{tabular}[c]{@{}c@{}}04/01 $\sim$15\\ (2016)\end{tabular} \\
\# Timestamps       & 336                                                              & 336                                                             & 336                                                             \\
Spatial granularity & \begin{tabular}[c]{@{}c@{}}Community \\ Area\end{tabular}        & \begin{tabular}[c]{@{}c@{}}Census \\ Tract\end{tabular}         & \begin{tabular}[c]{@{}c@{}}Census \\ Tract\end{tabular}         \\ \hline
\end{tabular}
\end{table}

\subsection{Baselines}
We compared MobiCLR with the following six baseline methods:  \textbf{(1) GAT \cite{velikovi2017graph}}, which is a GNN model that aims to train a two-week static OD flow; \textbf{(2) $x^o$}, which is the raw hourly count of outbound trips; \textbf{(3) $x^i$}, which is the raw hourly count of inbound trips; \textbf{(4) Mixing-up \cite{wickstrom2022mixing}}, which is a time series representation learning model that aims to predict the mixing proportion of two time series samples; \textbf{(5) TS-TCC \cite{tstcc}}, which is a time series representation learning model that uses a transformer-based autoregressive model to capture contextual information; and \textbf{(6) TS2vec \cite{Yue2021TS2VecTU}}, which is a time series representation learning model that incorporates contextual consistency through hierarchical contrastive loss. The reproduction details of the baselines can be found in the Appendix. 

\subsection{Prediction Results}
To evaluate the performance of the MobiCLR model, we followed the protocol presented in \cite{Yue2021TS2VecTU}. After the pre-training phase, the network parameters were fixed, and a ridge regression model was implemented at the top of the network. Subsequently, the encoders$ f_\theta^i$ and $f_\theta^o$ were discarded, and only $f_\theta^{io}$ was utilized for downstream applications. To obtain region-level representations, the pooling of $h_i^{io}$ across all timestamps was employed, resulting in $h_{i,\ast}^{io}$, which serves as the input for linear regression model. The regularization term of the linear regression model was selected through a grid search of \{0.1, 0.2, 0.5, 1, 2, 5, 10\}. The model's prediction performance for both single (educational attainment and income) and composite (social vulnerability index) index was evaluated using $R^2$. 

Table \ref{tab:prediction} presents the regression results for downstream applications in Chicago, New York, and Washington D.C. The results demonstrated that MobiCLR outperformed non-neural network models and state-of-the-art unsupervised methods across all downstream tasks in the three cities. It was observed that the performance of the non-neural network models were relatively poor compared to the neural networks models. The results also highlighted the disparate predictive capabilities of $x^o$ and $x^i$. This implies that semantic differences exist between inbound and outbound trips. Further, the GAT model, trained on aggregated OD flow information without considering temporal dynamics, exhibited lower performance than the time series representation learning models in most cases. This observation suggests that incorporating temporal dynamics into mobility data is effective. Additionally, it is worth noting that predicting the social vulnerability index, which account for the diverse elements of the urban environment, proved to be particularly challenging. However, MobiCLR demonstrates substantial performance gains in predicting social vulnerability across diverse urban settings, with improvements of 12.54\% in Chicago, 31.56\% in Washington D.C., and 3.13\% in New York over the runner-up models.

\begin{table*}[ht]
\centering
\caption{Regression results for the prediction of single and composite indicators. A simple linear regression model is trained on the training split with the pre-trained model, and evaluation metrics are reported for the regression on the test split. The best performance is noted in bold, and the runner-up in italics.
\\
* (EDU: Educational attainment, INC: Income, SVI: Social vulnerability index)
}
\label{tab:prediction}
\resizebox{\textwidth}{!}{
\begin{tabular}{ccccccccc}
\hline
\multicolumn{1}{l}{} & \multicolumn{1}{l}{} & $x_o$   & $x_i$   & GAT                          & Mixing-up                     & TS-TCC                       & TS2Vec                       & MobiCLR                      \\ \hline
\multirow{3}{*}{CH}  & EDU                  & -9.788  & -18.516 & 0.613 $\pm$ 0.070 & 0.698 $\pm$ 0.011 & 0.074 $\pm$ 0.298 & \textit{0.746 $\pm$ 0.020}  & \textbf{0.759 $\pm$ 0.020} \\
                     & INC                  & -1.279  & 0.184   & 0.398 $\pm$ 0.086 & \textit{0.776 $\pm$ 0.003}  & 0.520 $\pm$ 0.048 & 0.681 $\pm$ 0.066 & \textbf{0.785 $\pm$ 0.007} \\
                     & SVI                  & -13.731 & -11.181 & 0.312 $\pm$ 0.094 & \textit{0.543 $\pm$ 0.061}  & 0.194 $\pm$ 0.100 & 0.495 $\pm$ 0.044 & \textbf{0.611 $\pm$ 0.005} \\ \hline
\multirow{3}{*}{DC}  & EDU                  & -1.008  & -1.203  & 0.495 $\pm$ 0.044 & 0.442 $\pm$ 0.020  & 0.171 $\pm$ 0.023 & \textit{0.472 $\pm$ 0.017} & \textbf{0.601 $\pm$ 0.019} \\
                     & INC                  & -0.638  & -0.468  & 0.327 $\pm$ 0.132 & 0.292 $\pm$ 0.007  & 0.141 $\pm$ 0.010 & \textit{0.339 $\pm$ 0.017} & \textbf{0.416 $\pm$ 0.021} \\
                     & SVI                  & -1.705  & -1.644  & 0.198 $\pm$ 0.044 & \textit{0.363 $\pm$ 0.021}  & 0.069 $\pm$ 0.021 & 0.277 $\pm$ 0.017 & \textbf{0.446 $\pm$ 0.013} \\ \hline
\multirow{3}{*}{NY}  & EDU                  & -0.295  & -1.259  & 0.410 $\pm$ 0.027 & 0.736 $\pm$ 0.004  & 0.419 $\pm$ 0.012 & \textit{0.789 $\pm$ 0.017} & \textbf{0.790 $\pm$ 0.015} \\
                     & INC                  & 0.011   & -0.106  & 0.516 $\pm$ 0.027 & 0.734 $\pm$ 0.008  & 0.440 $\pm$ 0.014 & \textit{0.804 $\pm$ 0.012} & \textbf{0.817 $\pm$ 0.015} \\
                     & SVI                  & 1.295   & -0.714  & 0.417 $\pm$ 0.020 & 0.768 $\pm$ 0.004  & 0.417 $\pm$ 0.013 & \textit{0.785 $\pm$ 0.017} & \textbf{0.792 $\pm$ 0.009} \\ \hline
\end{tabular}
}
\end{table*}

\section{Analysis}
This Section assesses the choice of data augmentation strategy, conducts ablation studies to determine the significance of the model components, and explores the transferability of the knowledge learned across different cities.
\subsection{Data Augmentation}
\begin{figure*}[hbt!]

     \begin{subfigure}[h]{0.24\textwidth}
         \includegraphics[width=\textwidth]{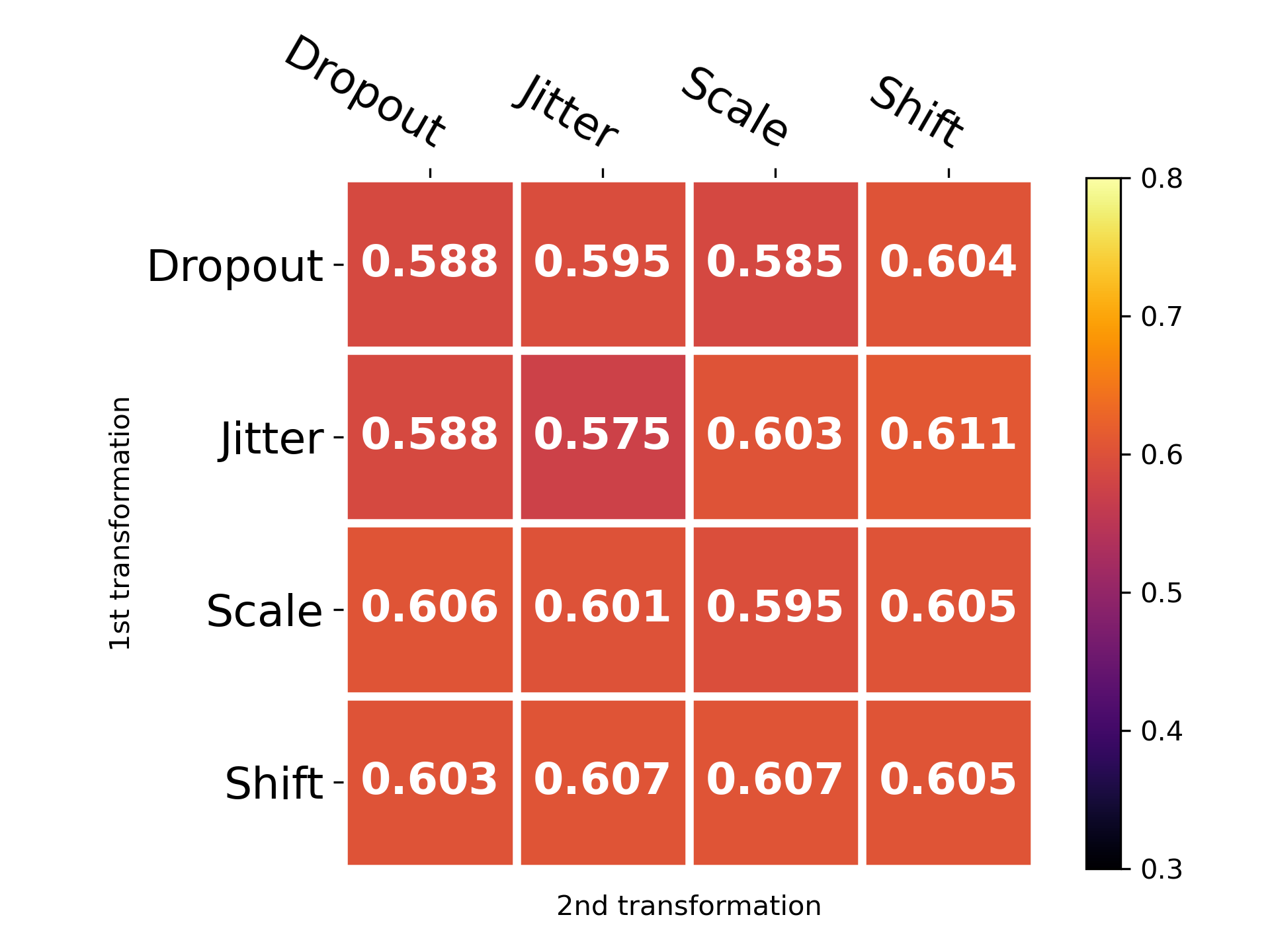}
         \caption[]{Chicago} 
         \label{fig:a}
     \end{subfigure}
     \begin{subfigure}[h]{0.24\textwidth}
         \includegraphics[width=\textwidth]{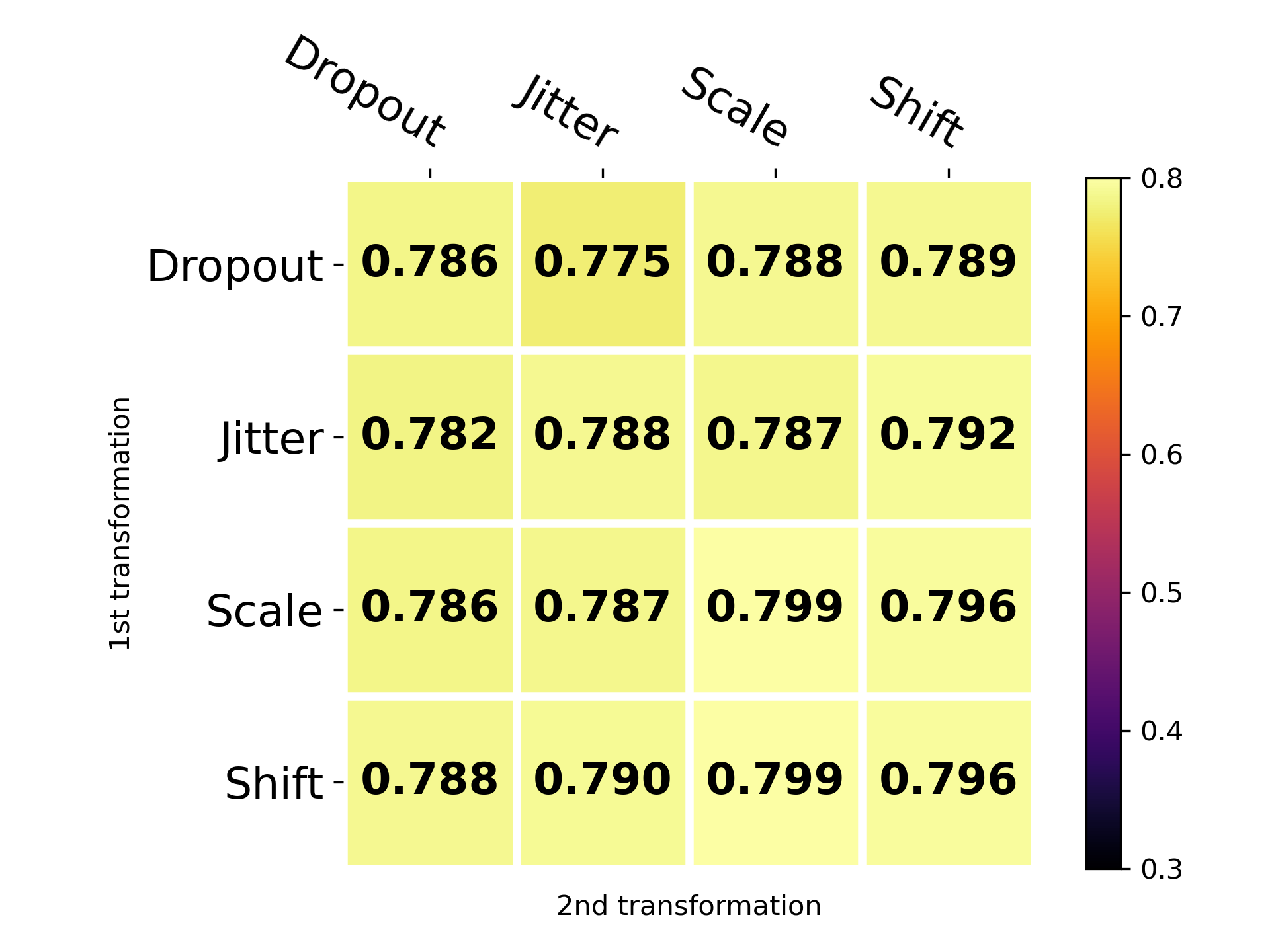}
         \caption[]{New York} 
         \label{fig:b}
     \end{subfigure}
     \begin{subfigure}[h]{0.24\textwidth}
         \includegraphics[width=\textwidth]{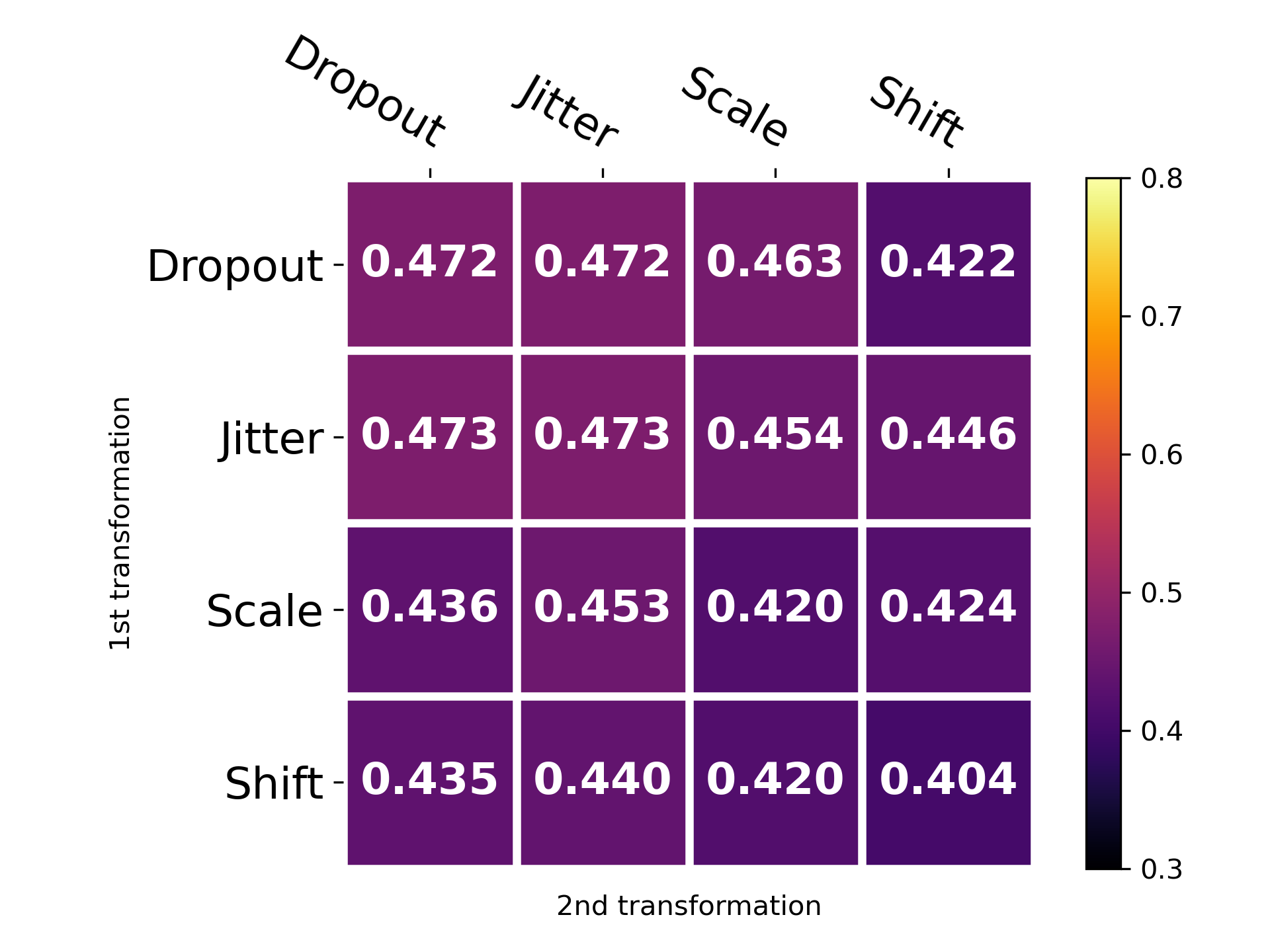}
         \caption[]{Washington DC} 
         \label{fig:c}
     \end{subfigure}
     \begin{subfigure}[h]{0.24\textwidth}
         \centering
         \includegraphics[width=\textwidth]{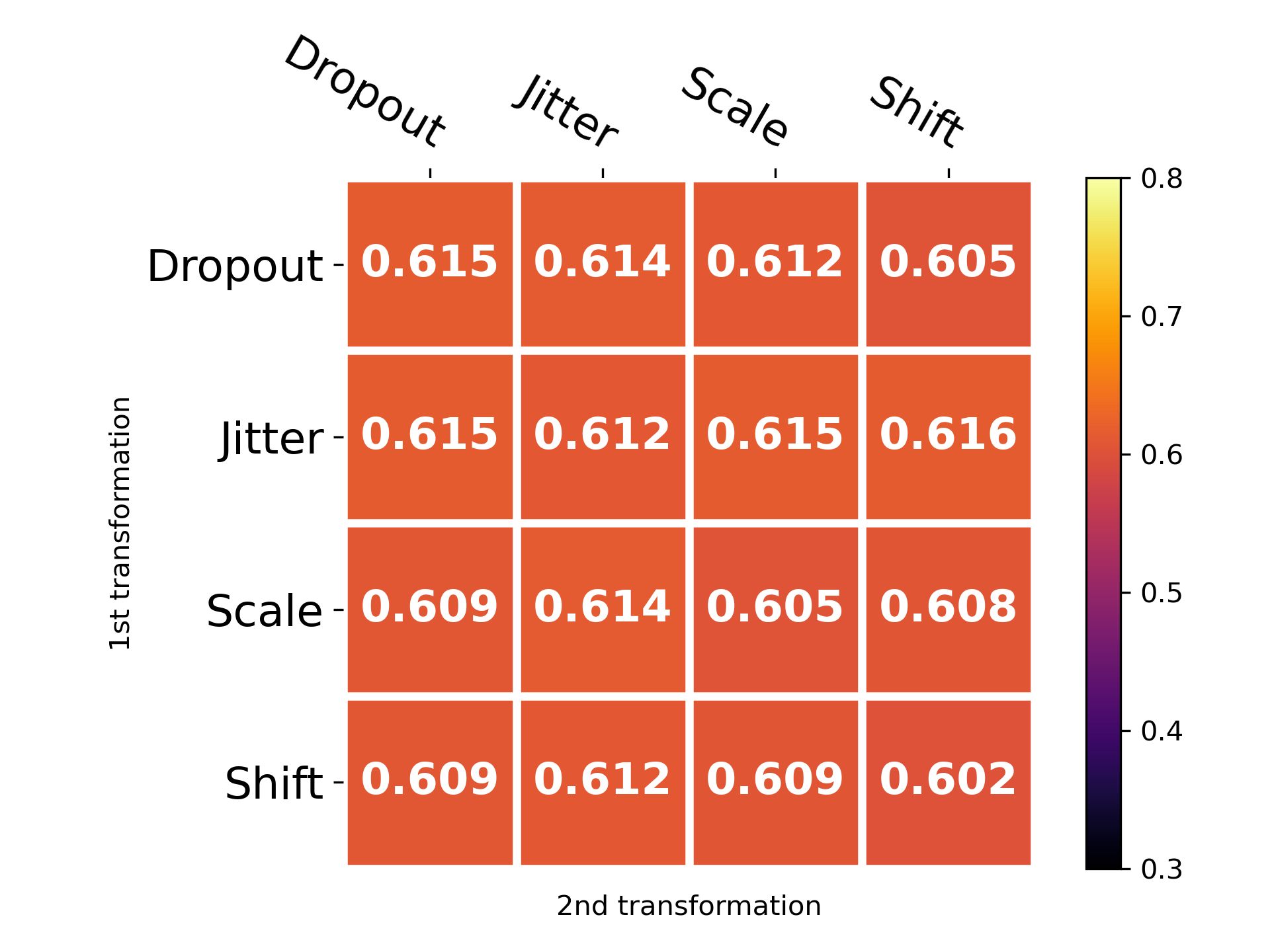}
         \caption[]{Average} 
         \label{fig:d}
     \end{subfigure}
        \caption{Results of social vulnerability prediction under different time series data augmentation strategies, including individual transformations and sequential compositions of two transformations. The diagonal entries represent the results of single transformations, while the off-diagonal entries show the results of compositions of two transformations.}
        \label{fig:aug}
\end{figure*}

To analyze the impact of data augmentation systematically, several common augmentations, including scaling, jittering, shift, and dropout, were considered. The objective was to evaluate the performance of the framework when the augmentations were applied individually or in pairs. 
Figure \ref{fig:aug} presents the social vulnerability index prediction results obtained by the framework for the three cities, showing the outcomes using individual transformations and the compositions of the transformations, along with their corresponding average scores.

Our investigation revealed that there existed no universal augmentation strategy to achieve optimal prediction results for all cities, as demonstrated by the results shown in Figure \ref{fig:a} to \ref{fig:c}. The effectiveness of different data augmentation techniques varied depending on the specific dataset. Each dataset in the three cities may have unique characteristics and patterns, which may require different augmentation strategies to improve model's performance. Nonetheless, a data augmentation strategy that includes shift after the jitter consistently produced robust results across all cities, as  evidenced by the average $R^2$ of social vulnerability prediction for the three cities presented in Figure \ref{fig:d}.

\subsection{Ablation Study}
The proposed model employs a joint learning objective, aiming to capture both inbound/outbound specific semantics and their correspondence. The hypothesis behind this approach suggests that combining these aspects would enhance representation learning compared to individual objectives. Specifically, $\mathcal{L}^i$ and $\mathcal{L}^o$ capture inbound and outbound-specific semantics, while $\mathcal{L}^{a}$ aligns the output features with these semantics for a more comprehensive representation. Ablation studies were conducted on the loss components to empirically validate this hypothesis.

As shown in Table \ref{tab:ablation}, our joint learning objective significantly improved social vulnerability index prediction accuracy. Through learning objective $\mathcal{L}^{a}$, the proposed model is capable of capturing both inbound and outbound mobility patterns, enhancing performance by producing a comprehensive representation of the underlying mobility patterns. Additionally, incorporating $\mathcal{L}^{a}$ may expand the model’s learning capacity by exposing it to a broader range of positive and negative sample variations, which could increase learning complexity and contribute to improved prediction results.

\begin{table}[]

\centering
\caption{Ablation studies on the loss components of MobiCLR, with the $R^2$ values of social vulnerability prediction.}
\label{tab:ablation}
\begin{tabular}{ccccc}
\hline
Loss                           & Chicago & New York & Washington D.C. & Avg \\ \hline
(1) $\mathcal{L}^o $ only                     & 0.610    & 0.790     & 0.143  & 0.513     \\
(2) $\mathcal{L}^i$ only                    & 0.590    & 0.738     & 0.350    &  0.559    \\
(3) w/o $\mathcal{L}^{a}$ & 0.604    & 0.781     & 0.310    &  0.565    \\
(4) $Full model$                                    & 0.611    & 0.792     & 0.446   &  0.616     \\ \hline
\end{tabular}
\end{table}

\subsection{Sensitivity Analysis}
\begin{figure*}[hbt!]
\centering
     \begin{subfigure}[h]{0.45\textwidth}
         \includegraphics[width=\textwidth]{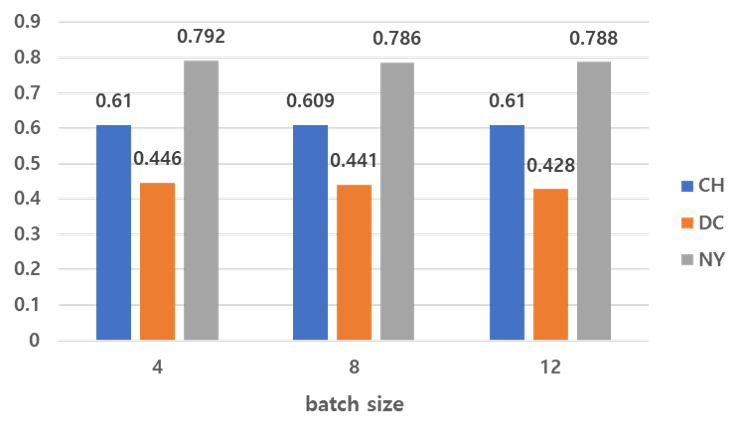}
         \caption[]{Batch size} 
         \label{fig:sena}
     \end{subfigure}
     \begin{subfigure}[h]{0.45\textwidth}
         \includegraphics[width=\textwidth]{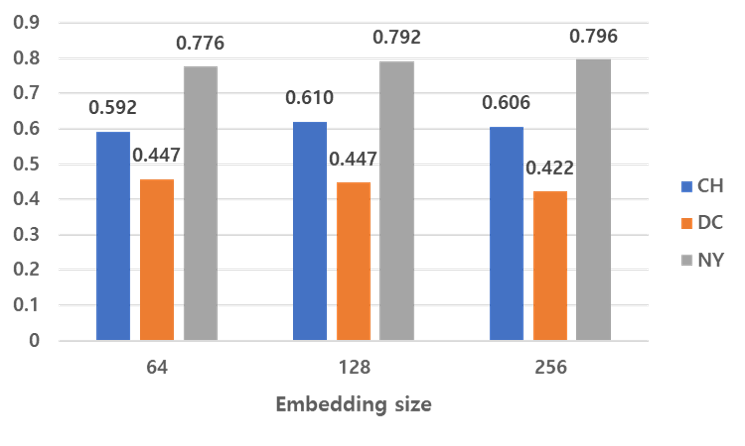}
         \caption[]{Embedding size} 
         \label{fig:senb}
     \end{subfigure}
        \caption{Sensitivity analysis}
        \label{fig:sen}
\end{figure*}

We investigate the robustness of the proposed model. To begin, we analyze the impact of three parameters: batch size (4, 8, 12) and embedding size (64, 128, 256) on the model's performance. Figure \ref{fig:sen} presents the performance of the proposed model corresponding to different values of. In general, the results indicate that the model shows the improved prediction result when batch size is set to 4, and embedding dimension is set to 128. The proposed model outperforms state-of-the-art baselines across most settings, with the exception of the social vulnerability index prediction result in New York when the embedding size is 64.

\subsection{Transferability Test}

A transferability test was conducted to assess the applicability of the learned model parameters from the source dataset to the target dataset. A model trained on mobility data from one city was utilized to predict social vulnerability in another city. The encoder of the model $f_\theta^{io}$ was kept fixed, and only the linear regressor was fine-tuned. 

\begin{figure}[hbt!]
\centering
         \includegraphics[width=0.43\textwidth]{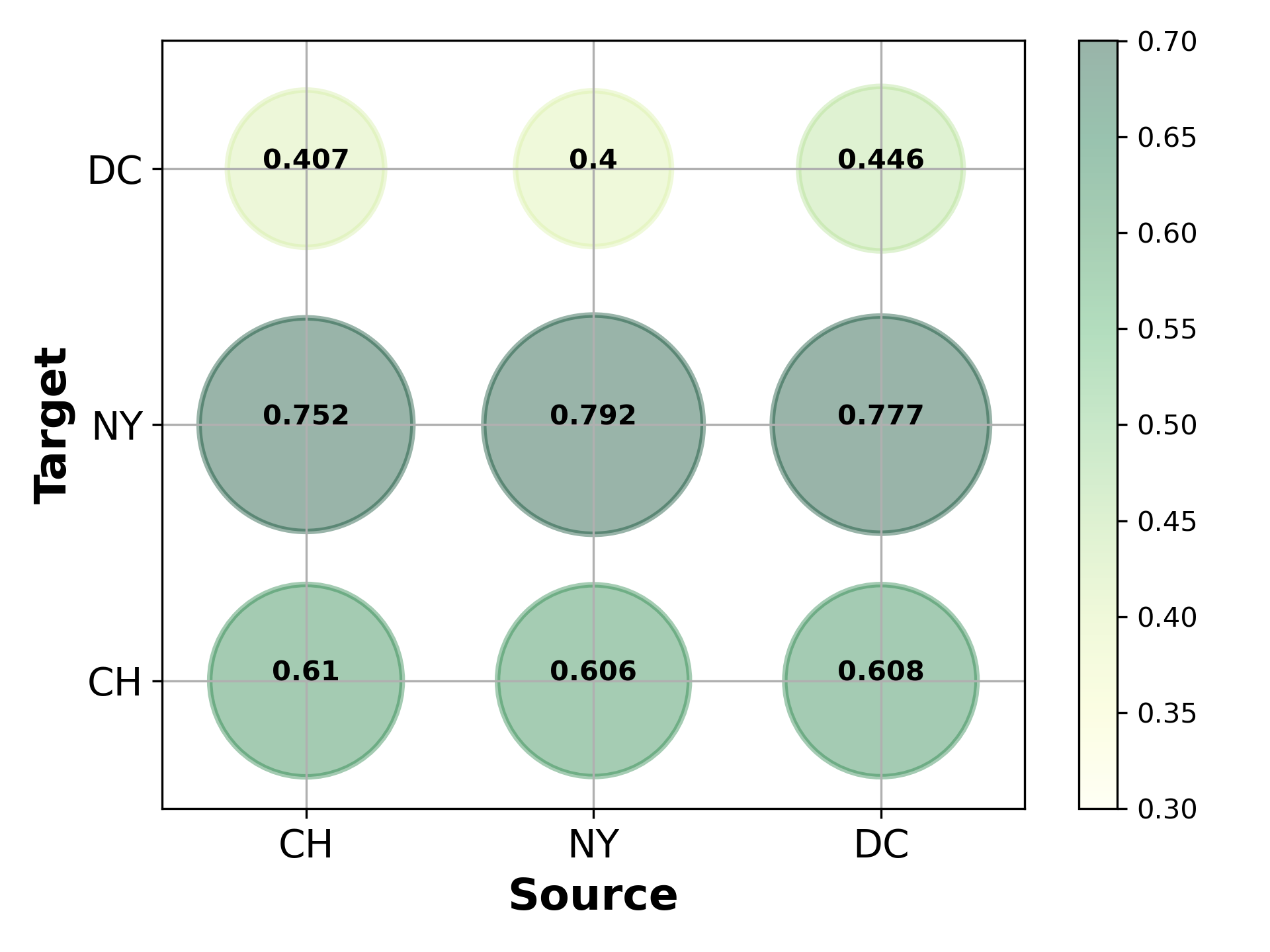}
        \caption{$R^2$ for the transferability test on social vulnerability index. The model is trained on one source city and then evaluated on other target city.}
        \label{fig:transferability}
\end{figure}

Experiments were conducted by varying the source-target city pairs to evaluate the social vulnerability prediction performance, as shown in Figure \ref{fig:transferability}. The diagonal line in the figure shows the highest correlation because it represents the same city transferred from the source to the target. Transfer experiments demonstrated  competitive performances compared to the best baseline. For example, in the "New York → Washington D.C." and "Washington D.C. → Chicago" scenarios, our model achieved an $R^2$ of 0.400 and 0.608, respectively, outperforming the second best approach in the non-transfer setting. These results validate our model's transferability for predicting social vulnerability across different cities, highlighting its broader applicability. 

Recent studies have revealed that vulnerable communities experience extended recovery periods, exacerbating the hardships \cite{flanagan2011social, dasgupta2020association}. These findings highlight the importance of identifying socially vulnerable areas, which are now recognized as critical components in disaster planning and mitigation strategies. In this context, the result of transferability test suggests that the proposed approach can be effectively applied in regions where social vulnerability has not yet been assessed, providing valuable insights for proactive responses to future natural hazards or external stressors.

\section{Conclusion}
This study proposed a novel method for learning region representations using the hourly counts of inbound and outbound trips. The proposed method is based on time series representation learning, designed to uncover the intrinsic characteristics embedded within dynamically changing inflow and outflow mobility patterns. Specifically, the model learns separate representations for inflow and outflow patterns and aligns them to create a unified view of mobility dynamics, allowing for a comprehensive understanding of regional mobility patterns. To validate the proposed model, we conducted experiments to predict income, educational attainment, and social vulnerability index in three cities: Chicago, New York, and Washington D.C. Experimental results demonstrated that the proposed approach achieved accurate predictions of these indicators using only two weeks of mobility data. Furthermore, transferability tests indicated that our approach can provide valuable insights in areas where such indicators have not been assessed. 

While our study has effectively utilized mobility data to predict socio-economic indicators and social vulnerability, the use of taxi or ride-hailing data may not fully represent the entire human mobility. According to \cite{statista}, about 17\% of people use taxis or ride-hailing services for their daily commutes. This limited representation of transportation modes could introduce biases in our assessments. As we look ahead to future research endeavors, we aim to utilize various modes of transportation such as buses and subways. This expanded approach will lead to a more comprehensive understanding of human mobility, and consequently, we anticipate achieving better prediction results. 

\section*{Acknowledgment}
This work was supported by the National Research Foundation of Korea (NRF) Basic Research Lab Grant (No. 2021R1A4A1033486) and Midcareer Research Grant (No. 2020R1A2C2010200) by the South Korean government, as well as the KAIST Artificial Intelligence Graduate School Program (2019-0-00075).

\section*{Guidelines for Artificial Intelligence (AI)-Generated Text}
During the preparation of this work the authors used ChatGPT in order to refine the writing and check the grammar.

\clearpage
\section*{Appendix}

\subsection*{Data Augmentation}
The data augmentation strategies used in this study are as follows. 
\begin{itemize}
\item \textbf{Shift:} The time series data is shifted by a single random scalar value, obtained by sampling from a Gaussian distribution with a mean of 0 and a standard deviation of 0.2 (i.e., $\epsilon\sim N(0,\ 0.2)$). Each time step is then shifted as ${\widetilde{x}}_t={\epsilon+x}_t$.
\item \textbf{Dropout:} It randomly drops (i.e., sets to zero) a certain proportion of the input data. This forces the model to learn more robust and generalized representations of the input data because it prevents the model from relying too heavily on any one particular feature or input. The dropout probability was set to 0.1.
\item \textbf{Jitter:} Independent and identically distributed Gaussian noise is added to each time step, sampled from a Gaussian distribution with a mean of 0 and a standard deviation of 0.2 (i.e., $\epsilon_t\sim N(0,\ 0.2)$). Each time step is now jittered as ${\widetilde{x}}_t={\epsilon_t}\times x_t$.
\item \textbf{Scale:} The time series data is scaled by a single random scalar value, obtained by sampling from a Gaussian distribution with a mean of 0 and a standard deviation of 0.2 (i.e., $\epsilon\sim N(0,\ 0.2)$). Each time step is then scaled as ${\widetilde{x}}_t={\epsilon\times x}_t$.
\end{itemize}

\subsection*{Details of Baselines}
\begin{itemize}
\item \textbf{GAT} \cite{velikovi2017graph} learns the static origin-destination flow matrix. Specifically, the model learns the conditional trip distribution using Kullback–Leibler divergence. The GAT model was configured with an embedding size of 48, and comprised two layers. The learning rate was set to 0.001 and optimization was performed using the Adam optimizer. 

\item \textbf{Inbound times series $(x^i)$} represents raw hourly counts of outbound trips. The inbound time series is normalized using the z-score, such that the set of observations has zero mean and unit variance. This results in a matrix $x^i\in R^{N\times T}$ and it was directly used for downstream applications.

\item \textbf{Outbound times series $(x^o)$} represents the raw hourly count of inbound trips and is normalized using z-score normalization. The resulting matrix $x^o\in R^{N\times T}$ was used directly in downstream applications.

\item \textbf{Mixing-up} \cite{wickstrom2022mixing} employs a novel augmentation and pretext task framework that aims to predict the mixing proportion of two time series samples. This approach generates new samples by considering a convex combination of two existing time series samples using a mixing parameter drawn from a beta distribution. The model used a modified version of the NT-Xent loss function to predict the mixing proportion. For implementation, the same beta distribution and model parameters as those reported in the original study were adopted. The open source code was obtained from “https://github.com/Wickstrom/ MixupContrastiveLearning”.

\item \textbf{TS-TCC} \cite{tstcc} utilizes a transformer-based autoregressive model to capture contextual information and ensure transferability. Their augmentation strategy comprises jitter and scale and permutation and jitter. TS-TCC imposes a challenging pretext task to learn the transformation-invariant information of a time series. The TS-TCC is designed to learn robust and discriminative features from time series data through a two-stage process. In the first stage, an input time series was passed through a temporal contrast module that enforces a challenging cross-view prediction task, resulting in the learning of contextual information. In the second stage, a contextual contrast module was applied to the learned representations to further enhance their discriminative power. For implementation, we referred to their configuration on Sleep EDF datasets. The open-source code was obtained from “https://github.com/emadeldeen24/TS-TCC”.

\item \textbf{TS2vec}\cite{Yue2021TS2VecTU} is a time series representation learning method that incorporates contextual consistency through a hierarchical contrastive loss. The method utilizes both instance-wise and temporal-wise contrastive losses to learn both instance discrimination and dynamic trends of time series data. Subseries sampling and masking were applied to augment the input time series. The hierarchical contrastive loss was computed by gradually pooling contextual representations along the time dimension and applying a loss function based on contextual consistency. The open-source code was obtained from “https://github.com/yuezhihan/ts2vec”.
\end{itemize}

 \bibliographystyle{cas-model2-names}

\bibliography{cas-refs}

\begin{thebibliography}{40}
\expandafter\ifx\csname natexlab\endcsname\relax\def\natexlab#1{#1}\fi
\providecommand{\url}[1]{\texttt{#1}}
\providecommand{\href}[2]{#2}
\providecommand{\path}[1]{#1}
\providecommand{\DOIprefix}{doi:}
\providecommand{\ArXivprefix}{arXiv:}
\providecommand{\URLprefix}{URL: }
\providecommand{\Pubmedprefix}{pmid:}
\providecommand{\doi}[1]{\href{http://dx.doi.org/#1}{\path{#1}}}
\providecommand{\Pubmed}[1]{\href{pmid:#1}{\path{#1}}}
\providecommand{\bibinfo}[2]{#2}
\ifx\xfnm\relax \def\xfnm[#1]{\unskip,\space#1}\fi
\bibitem[{Adger(2006)}]{adger2006vulnerability}
\bibinfo{author}{Adger, W.N.}, \bibinfo{year}{2006}.
\newblock \bibinfo{title}{Vulnerability}.
\newblock \bibinfo{journal}{Global environmental change} \bibinfo{volume}{16}, \bibinfo{pages}{268--281}.
\bibitem[{Aiken et~al.(2022)Aiken, Bellue, Karlan, Udry and Blumenstock}]{aiken2022machine}
\bibinfo{author}{Aiken, E.}, \bibinfo{author}{Bellue, S.}, \bibinfo{author}{Karlan, D.}, \bibinfo{author}{Udry, C.}, \bibinfo{author}{Blumenstock, J.E.}, \bibinfo{year}{2022}.
\newblock \bibinfo{title}{Machine learning and phone data can improve targeting of humanitarian aid}.
\newblock \bibinfo{journal}{Nature} \bibinfo{volume}{603}, \bibinfo{pages}{864--870}.
\bibitem[{Bankoff et~al.(2004)Bankoff, Frerks, Hilhorst and Hilhorst}]{bankoff2004mapping}
\bibinfo{author}{Bankoff, G.}, \bibinfo{author}{Frerks, G.}, \bibinfo{author}{Hilhorst, T.}, \bibinfo{author}{Hilhorst, D.}, \bibinfo{year}{2004}.
\newblock \bibinfo{title}{Mapping vulnerability: disasters, development, and people}.
\newblock \bibinfo{publisher}{Routledge}.
\bibitem[{Carroll and Prentice(2021)}]{carroll2021community}
\bibinfo{author}{Carroll, R.}, \bibinfo{author}{Prentice, C.R.}, \bibinfo{year}{2021}.
\newblock \bibinfo{title}{Community vulnerability and mobility: What matters most in spatio-temporal modeling of the covid-19 pandemic?}
\newblock \bibinfo{journal}{Social Science \& Medicine} \bibinfo{volume}{287}, \bibinfo{pages}{114395}.
\bibitem[{Chen et~al.(2021)Chen, Zou, Li, Li, Yang and Chen}]{chen2021multiple}
\bibinfo{author}{Chen, Y.}, \bibinfo{author}{Zou, X.}, \bibinfo{author}{Li, K.}, \bibinfo{author}{Li, K.}, \bibinfo{author}{Yang, X.}, \bibinfo{author}{Chen, C.}, \bibinfo{year}{2021}.
\newblock \bibinfo{title}{Multiple local 3d cnns for region-based prediction in smart cities}.
\newblock \bibinfo{journal}{Information Sciences} \bibinfo{volume}{542}, \bibinfo{pages}{476--491}.
\bibitem[{Chi et~al.(2022)Chi, Fang, Chatterjee and Blumenstock}]{chi2022microestimates}
\bibinfo{author}{Chi, G.}, \bibinfo{author}{Fang, H.}, \bibinfo{author}{Chatterjee, S.}, \bibinfo{author}{Blumenstock, J.E.}, \bibinfo{year}{2022}.
\newblock \bibinfo{title}{Microestimates of wealth for all low-and middle-income countries}.
\newblock \bibinfo{journal}{Proceedings of the National Academy of Sciences} \bibinfo{volume}{119}, \bibinfo{pages}{e2113658119}.
\bibitem[{Cutter et~al.(2003)Cutter, Boruff and Shirley}]{cutter2003social}
\bibinfo{author}{Cutter, S.L.}, \bibinfo{author}{Boruff, B.J.}, \bibinfo{author}{Shirley, W.L.}, \bibinfo{year}{2003}.
\newblock \bibinfo{title}{Social vulnerability to environmental hazards}.
\newblock \bibinfo{journal}{Social science quarterly} \bibinfo{volume}{84}, \bibinfo{pages}{242--261}.
\bibitem[{Dasgupta et~al.(2020)Dasgupta, Bowen, Leidner, Fletcher, Musial, Rose, Cha, Kang, Dirlikov, Pevzner et~al.}]{dasgupta2020association}
\bibinfo{author}{Dasgupta, S.}, \bibinfo{author}{Bowen, V.B.}, \bibinfo{author}{Leidner, A.}, \bibinfo{author}{Fletcher, K.}, \bibinfo{author}{Musial, T.}, \bibinfo{author}{Rose, C.}, \bibinfo{author}{Cha, A.}, \bibinfo{author}{Kang, G.}, \bibinfo{author}{Dirlikov, E.}, \bibinfo{author}{Pevzner, E.}, et~al., \bibinfo{year}{2020}.
\newblock \bibinfo{title}{Association between social vulnerability and a county’s risk for becoming a covid-19 hotspot—united states, june 1--july 25, 2020}.
\newblock \bibinfo{journal}{Morbidity and Mortality Weekly Report} \bibinfo{volume}{69}, \bibinfo{pages}{1535}.
\bibitem[{Eldele et~al.(2021)Eldele, Ragab, Chen, Wu, Kwoh, Li and Guan}]{tstcc}
\bibinfo{author}{Eldele, E.}, \bibinfo{author}{Ragab, M.}, \bibinfo{author}{Chen, Z.}, \bibinfo{author}{Wu, M.}, \bibinfo{author}{Kwoh, C.K.}, \bibinfo{author}{Li, X.}, \bibinfo{author}{Guan, C.}, \bibinfo{year}{2021}.
\newblock \bibinfo{title}{Time-series representation learning via temporal and contextual contrasting}, in: \bibinfo{booktitle}{International Joint Conference on Artificial Intelligence}.
\bibitem[{Flanagan et~al.(2011)Flanagan, Gregory, Hallisey, Heitgerd and Lewis}]{flanagan2011social}
\bibinfo{author}{Flanagan, B.E.}, \bibinfo{author}{Gregory, E.W.}, \bibinfo{author}{Hallisey, E.J.}, \bibinfo{author}{Heitgerd, J.L.}, \bibinfo{author}{Lewis, B.}, \bibinfo{year}{2011}.
\newblock \bibinfo{title}{A social vulnerability index for disaster management}.
\newblock \bibinfo{journal}{Journal of homeland security and emergency management} \bibinfo{volume}{8}, \bibinfo{pages}{0000102202154773551792}.
\bibitem[{Franceschi et~al.(2019)Franceschi, Dieuleveut and Jaggi}]{franceschi2019unsupervised}
\bibinfo{author}{Franceschi, J.Y.}, \bibinfo{author}{Dieuleveut, A.}, \bibinfo{author}{Jaggi, M.}, \bibinfo{year}{2019}.
\newblock \bibinfo{title}{Unsupervised scalable representation learning for multivariate time series}.
\newblock \bibinfo{journal}{Advances in neural information processing systems} \bibinfo{volume}{32}.
\bibitem[{Fu et~al.(2019)Fu, Wang, Du, Wu and Li}]{fu2019efficient}
\bibinfo{author}{Fu, Y.}, \bibinfo{author}{Wang, P.}, \bibinfo{author}{Du, J.}, \bibinfo{author}{Wu, L.}, \bibinfo{author}{Li, X.}, \bibinfo{year}{2019}.
\newblock \bibinfo{title}{Efficient region embedding with multi-view spatial networks: A perspective of locality-constrained spatial autocorrelations}, in: \bibinfo{booktitle}{Proceedings of the AAAI Conference on Artificial Intelligence}, pp. \bibinfo{pages}{906--913}.
\bibitem[{Gao et~al.(2020)Gao, Zhang, Huang, Yin, Yang and Shao}]{gao2020semantic}
\bibinfo{author}{Gao, C.}, \bibinfo{author}{Zhang, Z.}, \bibinfo{author}{Huang, C.}, \bibinfo{author}{Yin, H.}, \bibinfo{author}{Yang, Q.}, \bibinfo{author}{Shao, J.}, \bibinfo{year}{2020}.
\newblock \bibinfo{title}{Semantic trajectory representation and retrieval via hierarchical embedding}.
\newblock \bibinfo{journal}{Information Sciences} \bibinfo{volume}{538}, \bibinfo{pages}{176--192}.
\bibitem[{Huang et~al.(2021)Huang, Wang, Sheng, Ng and Rajagopal}]{huang2021m3g}
\bibinfo{author}{Huang, T.}, \bibinfo{author}{Wang, Z.}, \bibinfo{author}{Sheng, H.}, \bibinfo{author}{Ng, A.Y.}, \bibinfo{author}{Rajagopal, R.}, \bibinfo{year}{2021}.
\newblock \bibinfo{title}{M3g: Learning urban neighborhood representation from multi-modal multi-graph}, in: \bibinfo{booktitle}{Proceedings of the DeepSpatial 2021: 2nd ACM KDD Workshop on Deep Learning for Spatio-Temporal Data, Applications and Systems}.
\bibitem[{Kim and Yoon(2022)}]{kim2022effective}
\bibinfo{author}{Kim, N.}, \bibinfo{author}{Yoon, Y.}, \bibinfo{year}{2022}.
\newblock \bibinfo{title}{Effective urban region representation learning using heterogeneous urban graph attention network (hugat)}.
\newblock \bibinfo{journal}{arXiv preprint arXiv:2202.09021} .
\bibitem[{Li et~al.(2023)Li, Huang, Cong, Wang and Wang}]{li2023urban}
\bibinfo{author}{Li, Y.}, \bibinfo{author}{Huang, W.}, \bibinfo{author}{Cong, G.}, \bibinfo{author}{Wang, H.}, \bibinfo{author}{Wang, Z.}, \bibinfo{year}{2023}.
\newblock \bibinfo{title}{Urban region representation learning with openstreetmap building footprints}, in: \bibinfo{booktitle}{Proceedings of the 29th ACM SIGKDD Conference on Knowledge Discovery and Data Mining}, pp. \bibinfo{pages}{1363--1373}.
\bibitem[{Li et~al.(2024)Li, Huang, Zhao, Yang, Gong and Chen}]{li2024urban}
\bibinfo{author}{Li, Z.}, \bibinfo{author}{Huang, W.}, \bibinfo{author}{Zhao, K.}, \bibinfo{author}{Yang, M.}, \bibinfo{author}{Gong, Y.}, \bibinfo{author}{Chen, M.}, \bibinfo{year}{2024}.
\newblock \bibinfo{title}{Urban region embedding via multi-view contrastive prediction}, in: \bibinfo{booktitle}{Proceedings of the AAAI Conference on Artificial Intelligence}, pp. \bibinfo{pages}{8724--8732}.
\bibitem[{Liang et~al.(2023)Liang, Zhao and Biljecki}]{liang2023revealing}
\bibinfo{author}{Liang, X.}, \bibinfo{author}{Zhao, T.}, \bibinfo{author}{Biljecki, F.}, \bibinfo{year}{2023}.
\newblock \bibinfo{title}{Revealing spatio-temporal evolution of urban visual environments with street view imagery}.
\newblock \bibinfo{journal}{Landscape and Urban Planning} \bibinfo{volume}{237}, \bibinfo{pages}{104802}.
\bibitem[{Liu et~al.(2024)Liu, Zhang, Zhu, Guan and Kwong}]{liu2024exploring}
\bibinfo{author}{Liu, C.}, \bibinfo{author}{Zhang, H.}, \bibinfo{author}{Zhu, G.}, \bibinfo{author}{Guan, H.}, \bibinfo{author}{Kwong, S.}, \bibinfo{year}{2024}.
\newblock \bibinfo{title}{Exploring trajectory embedding via spatial-temporal propagation for dynamic region representations}.
\newblock \bibinfo{journal}{Information Sciences} \bibinfo{volume}{668}, \bibinfo{pages}{120516}.
\bibitem[{Liu et~al.(2021)Liu, Li, Ji, Xie, Du, Teng and Zhang}]{liu2021urban}
\bibinfo{author}{Liu, J.}, \bibinfo{author}{Li, T.}, \bibinfo{author}{Ji, S.}, \bibinfo{author}{Xie, P.}, \bibinfo{author}{Du, S.}, \bibinfo{author}{Teng, F.}, \bibinfo{author}{Zhang, J.}, \bibinfo{year}{2021}.
\newblock \bibinfo{title}{Urban flow pattern mining based on multi-source heterogeneous data fusion and knowledge graph embedding}.
\newblock \bibinfo{journal}{IEEE Transactions on Knowledge and Data Engineering} \bibinfo{volume}{35}, \bibinfo{pages}{2133--2146}.
\bibitem[{Meng et~al.(2023)Meng, Qian, Liu, Cui, Xu and Shen}]{meng2023mhccl}
\bibinfo{author}{Meng, Q.}, \bibinfo{author}{Qian, H.}, \bibinfo{author}{Liu, Y.}, \bibinfo{author}{Cui, L.}, \bibinfo{author}{Xu, Y.}, \bibinfo{author}{Shen, Z.}, \bibinfo{year}{2023}.
\newblock \bibinfo{title}{Mhccl: Masked hierarchical cluster-wise contrastive learning for multivariate time series}, in: \bibinfo{booktitle}{Proceedings of the AAAI Conference on Artificial Intelligence}, pp. \bibinfo{pages}{9153--9161}.
\bibitem[{Oord et~al.(2016)Oord, Dieleman, Zen, Simonyan, Vinyals, Graves, Kalchbrenner, Senior and Kavukcuoglu}]{oord2016wavenet}
\bibinfo{author}{Oord, A.v.d.}, \bibinfo{author}{Dieleman, S.}, \bibinfo{author}{Zen, H.}, \bibinfo{author}{Simonyan, K.}, \bibinfo{author}{Vinyals, O.}, \bibinfo{author}{Graves, A.}, \bibinfo{author}{Kalchbrenner, N.}, \bibinfo{author}{Senior, A.}, \bibinfo{author}{Kavukcuoglu, K.}, \bibinfo{year}{2016}.
\newblock \bibinfo{title}{Wavenet: A generative model for raw audio}.
\newblock \bibinfo{journal}{arXiv preprint arXiv:1609.03499} .
\bibitem[{Oord et~al.(2018)Oord, Li and Vinyals}]{oord2018representation}
\bibinfo{author}{Oord, A.v.d.}, \bibinfo{author}{Li, Y.}, \bibinfo{author}{Vinyals, O.}, \bibinfo{year}{2018}.
\newblock \bibinfo{title}{Representation learning with contrastive predictive coding}.
\newblock \bibinfo{journal}{arXiv preprint arXiv:1807.03748} .
\bibitem[{Paszke et~al.(2019)Paszke, Gross, Massa, Lerer, Bradbury, Chanan, Killeen, Lin, Gimelshein, Antiga et~al.}]{paszke2019pytorch}
\bibinfo{author}{Paszke, A.}, \bibinfo{author}{Gross, S.}, \bibinfo{author}{Massa, F.}, \bibinfo{author}{Lerer, A.}, \bibinfo{author}{Bradbury, J.}, \bibinfo{author}{Chanan, G.}, \bibinfo{author}{Killeen, T.}, \bibinfo{author}{Lin, Z.}, \bibinfo{author}{Gimelshein, N.}, \bibinfo{author}{Antiga, L.}, et~al., \bibinfo{year}{2019}.
\newblock \bibinfo{title}{Pytorch: An imperative style, high-performance deep learning library}.
\newblock \bibinfo{journal}{Advances in neural information processing systems} \bibinfo{volume}{32}.
\bibitem[{Sohn(2016)}]{sohn2016improved}
\bibinfo{author}{Sohn, K.}, \bibinfo{year}{2016}.
\newblock \bibinfo{title}{Improved deep metric learning with multi-class n-pair loss objective}.
\newblock \bibinfo{journal}{Advances in neural information processing systems} \bibinfo{volume}{29}.
\bibitem[{Statista(2024)}]{statista}
\bibinfo{author}{Statista}, \bibinfo{year}{2024}.
\newblock \bibinfo{title}{Most common modes of transportation for commuting in the u.s. as of december 2023}.
\newblock \bibinfo{howpublished}{\url{https://www.statista.com/forecasts/997176/most-common-modes-of-transportation-for-commuting-in-the-us}}.
\bibitem[{Tonekaboni et~al.(2021)Tonekaboni, Eytan and Goldenberg}]{tonekaboni2020unsupervised}
\bibinfo{author}{Tonekaboni, S.}, \bibinfo{author}{Eytan, D.}, \bibinfo{author}{Goldenberg, A.}, \bibinfo{year}{2021}.
\newblock \bibinfo{title}{Unsupervised representation learning for time series with temporal neighborhood coding}, in: \bibinfo{booktitle}{International Conference on Learning Representations}.
\bibitem[{Veličković et~al.(2017)Veličković, Cucurull, Casanova, Romero, Liò and Bengio}]{velikovi2017graph}
\bibinfo{author}{Veličković, P.}, \bibinfo{author}{Cucurull, G.}, \bibinfo{author}{Casanova, A.}, \bibinfo{author}{Romero, A.}, \bibinfo{author}{Liò, P.}, \bibinfo{author}{Bengio, Y.}, \bibinfo{year}{2017}.
\newblock \bibinfo{title}{Graph attention networks}.
\newblock \bibinfo{journal}{International Conference on Learning Representations} .
\bibitem[{Wang and Li(2017)}]{wang2017region}
\bibinfo{author}{Wang, H.}, \bibinfo{author}{Li, Z.}, \bibinfo{year}{2017}.
\newblock \bibinfo{title}{Region representation learning via mobility flow}, in: \bibinfo{booktitle}{Proceedings of the 2017 ACM on Conference on Information and Knowledge Management}, pp. \bibinfo{pages}{237--246}.
\bibitem[{Wickstr{\o}m et~al.(2022)Wickstr{\o}m, Kampffmeyer, Mikalsen and Jenssen}]{wickstrom2022mixing}
\bibinfo{author}{Wickstr{\o}m, K.}, \bibinfo{author}{Kampffmeyer, M.}, \bibinfo{author}{Mikalsen, K.{\O}.}, \bibinfo{author}{Jenssen, R.}, \bibinfo{year}{2022}.
\newblock \bibinfo{title}{Mixing up contrastive learning: Self-supervised representation learning for time series}.
\newblock \bibinfo{journal}{Pattern Recognition Letters} \bibinfo{volume}{155}, \bibinfo{pages}{54--61}.
\bibitem[{Wu et~al.(2022)Wu, Yan, Fan, Pan, Zhu, Zheng, Cheng and Wang}]{wu2022multi}
\bibinfo{author}{Wu, S.}, \bibinfo{author}{Yan, X.}, \bibinfo{author}{Fan, X.}, \bibinfo{author}{Pan, S.}, \bibinfo{author}{Zhu, S.}, \bibinfo{author}{Zheng, C.}, \bibinfo{author}{Cheng, M.}, \bibinfo{author}{Wang, C.}, \bibinfo{year}{2022}.
\newblock \bibinfo{title}{Multi-graph fusion networks for urban region embedding}.
\newblock \bibinfo{journal}{arXiv preprint arXiv:2201.09760} .
\bibitem[{Wu et~al.(2018)Wu, Xiong, Yu and Lin}]{wu2018unsupervised}
\bibinfo{author}{Wu, Z.}, \bibinfo{author}{Xiong, Y.}, \bibinfo{author}{Yu, S.X.}, \bibinfo{author}{Lin, D.}, \bibinfo{year}{2018}.
\newblock \bibinfo{title}{Unsupervised feature learning via non-parametric instance discrimination}, in: \bibinfo{booktitle}{Proceedings of the IEEE conference on computer vision and pattern recognition}, pp. \bibinfo{pages}{3733--3742}.
\bibitem[{Xia et~al.(2023)Xia, Hu, Chi and Chen}]{xia2023assessing}
\bibinfo{author}{Xia, C.}, \bibinfo{author}{Hu, Y.}, \bibinfo{author}{Chi, G.}, \bibinfo{author}{Chen, J.}, \bibinfo{year}{2023}.
\newblock \bibinfo{title}{Assessing dynamics of human vulnerability at community level--using mobility data}.
\newblock \bibinfo{journal}{International Journal of Disaster Risk Reduction} \bibinfo{volume}{96}, \bibinfo{pages}{103964}.
\bibitem[{Yao et~al.(2018)Yao, Fu, Liu, Hu and Xiong}]{yao2018representing}
\bibinfo{author}{Yao, Z.}, \bibinfo{author}{Fu, Y.}, \bibinfo{author}{Liu, B.}, \bibinfo{author}{Hu, W.}, \bibinfo{author}{Xiong, H.}, \bibinfo{year}{2018}.
\newblock \bibinfo{title}{Representing urban functions through zone embedding with human mobility patterns}, in: \bibinfo{booktitle}{Proceedings of the Twenty-Seventh International Joint Conference on Artificial Intelligence (IJCAI-18)}.
\bibitem[{Yu and Koltun(2016)}]{Fisher}
\bibinfo{author}{Yu, F.}, \bibinfo{author}{Koltun, V.}, \bibinfo{year}{2016}.
\newblock \bibinfo{title}{Multi-scale context aggregation by dilated convolutions}, in: \bibinfo{booktitle}{International Conference on Learning Representations, San Juan, Puerto Rico, May 2-4, 2016, Conference Track Proceedings}.
\bibitem[{Yue et~al.(2022)Yue, Wang, Duan, Yang, Huang, Tong and Xu}]{Yue2021TS2VecTU}
\bibinfo{author}{Yue, Z.}, \bibinfo{author}{Wang, Y.}, \bibinfo{author}{Duan, J.}, \bibinfo{author}{Yang, T.}, \bibinfo{author}{Huang, C.}, \bibinfo{author}{Tong, Y.}, \bibinfo{author}{Xu, B.}, \bibinfo{year}{2022}.
\newblock \bibinfo{title}{Ts2vec: Towards universal representation of time series}, in: \bibinfo{booktitle}{AAAI Conference on Artificial Intelligence}.
\bibitem[{Zhang et~al.(2022)Zhang, Long and Cong}]{zhang2022region}
\bibinfo{author}{Zhang, L.}, \bibinfo{author}{Long, C.}, \bibinfo{author}{Cong, G.}, \bibinfo{year}{2022}.
\newblock \bibinfo{title}{Region embedding with intra and inter-view contrastive learning}.
\newblock \bibinfo{journal}{IEEE Transactions on Knowledge and Data Engineering} \bibinfo{volume}{35}, \bibinfo{pages}{9031--9036}.
\bibitem[{Zhang et~al.(2021a)Zhang, Li, Li and Hui}]{zhang2021multi}
\bibinfo{author}{Zhang, M.}, \bibinfo{author}{Li, T.}, \bibinfo{author}{Li, Y.}, \bibinfo{author}{Hui, P.}, \bibinfo{year}{2021}a.
\newblock \bibinfo{title}{Multi-view joint graph representation learning for urban region embedding}, in: \bibinfo{booktitle}{Proceedings of the twenty-ninth international conference on international joint conferences on artificial intelligence}, pp. \bibinfo{pages}{4431--4437}.
\bibitem[{Zhang et~al.(2021b)Zhang, Duan and Li}]{zhang2021unveiling}
\bibinfo{author}{Zhang, T.}, \bibinfo{author}{Duan, X.}, \bibinfo{author}{Li, Y.}, \bibinfo{year}{2021}b.
\newblock \bibinfo{title}{Unveiling transit mobility structure towards sustainable cities: An integrated graph embedding approach}.
\newblock \bibinfo{journal}{Sustainable Cities and Society} \bibinfo{volume}{72}, \bibinfo{pages}{103027}.
\bibitem[{Zhao et~al.(2023)Zhao, Qi, Trisedya, Su, Zhang and Ren}]{zhao2023learning}
\bibinfo{author}{Zhao, Y.}, \bibinfo{author}{Qi, J.}, \bibinfo{author}{Trisedya, B.D.}, \bibinfo{author}{Su, Y.}, \bibinfo{author}{Zhang, R.}, \bibinfo{author}{Ren, H.}, \bibinfo{year}{2023}.
\newblock \bibinfo{title}{Learning region similarities via graph-based deep metric learning}.
\newblock \bibinfo{journal}{IEEE Transactions on Knowledge and Data Engineering} \bibinfo{volume}{35}, \bibinfo{pages}{10237--10250}.

\end{thebibliography}


\end{document}